\documentclass{article}

\PassOptionsToPackage{numbers, compress}{natbib}

   \usepackage[final]{neurips_2024}

\usepackage[utf8]{inputenc} 
\usepackage[T1]{fontenc}    
\usepackage{hyperref}       
\usepackage{cleveref}
\usepackage{url}            
\usepackage{booktabs}       
\usepackage{amsfonts}       
\usepackage{nicefrac}       
\usepackage{microtype}      
\usepackage{xcolor}         
\usepackage{graphicx} 
\usepackage{subcaption}

\author{
  Keyu Wang \\
  Mila, Quebec AI Institute \\
  McGill University\\
  \And
  Abdullah Norozi Iranzad
  \And
  Scott Schaffter\\
  Google\\
  \And 
  Meg Risdal\\
  Google\\
  \And
  Doina Precup \\
  Mila, Quebec AI Institute \\
  McGill University\\
  \And
  Jonathan Lebensold \\
  Mila, Quebec AI Institute \\
  McGill University\\
}

\title{Mitigating Downstream Model Risks via Model Provenance}
\begin{document}

\maketitle
 

\begin{abstract}
Research and industry are rapidly advancing the innovation and adoption of foundation model-based systems, yet the tools for managing these models have not kept pace. Understanding the provenance and lineage of models is critical for researchers, industry, regulators, and public trust. While model cards \citep{mitchell2019model} and system cards \citep{gursoy2022system} were designed to provide transparency, they fall short in key areas: tracing model genealogy, enabling machine readability, offering reliable centralized management systems, and fostering consistent creation incentives. This challenge mirrors issues in software supply chain security, but AI/ML remains at an earlier stage of maturity. Addressing these gaps requires industry-standard tooling that can be adopted by foundation model publishers, open-source model innovators, and major distribution platforms. We propose a machine-readable model specification format to simplify the creation of model records, thereby reducing error-prone human effort, notably when a new model inherits most of its design from a foundation model. Our solution explicitly traces relationships between upstream and downstream models, enhancing transparency and traceability across the model lifecycle. To facilitate the adoption, we introduce the \textit{unified model record} (UMR) repository, a semantically versioned system that automates the publication of model records to multiple formats (PDF, HTML, LaTeX) and provides a hosted web interface. This proof of concept aims to set a new standard for managing foundation models, bridging the gap between innovation and responsible model management.\end{abstract}

\section{Introduction}

Despite the incredible performance of frontier and foundation models \citep{bommasani2021opportunities} and their increasing use in production applications \citep{kawaharazuka2024real}, the lack of transparency regarding the datasets and upstream models used during training remains a critical issue \citep{bommasani2023ecosystem}. Without clear documentation, downstream developers risk inheriting harmful biases or flaws from upstream models, sometimes surfacing months or even years after a model's release \citep{kapoor2024position}. However, our goal is not to compel model publishers—especially those offering closed-source, API-based models like GPT-4 \citep{openai2024gpt4}—to disclose proprietary training details. Instead, we propose an open-source contribution addressing the following question:

\begin{center}
    \textit{How do we warn downstream model providers of upstream risks?}
\end{center}

\begin{figure}
    \centering
    \includegraphics[width=1\textwidth]{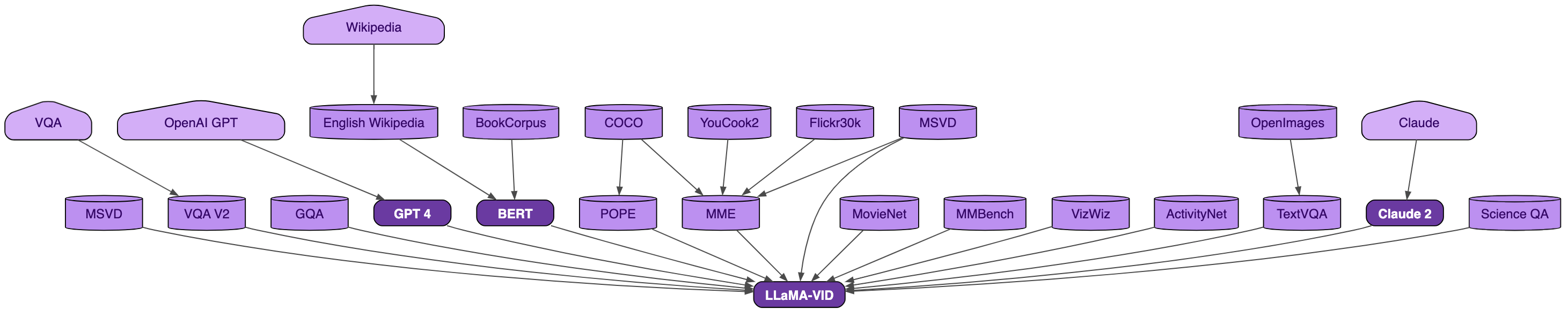}
    \caption{The provenance graph for the Llama-VID Short Video, a video-to-text captioning model built using a number of open-source foundation models.}
    \label{fig:complex_provenance}
\end{figure}

Increasingly, foundation models rely on Internet-scale datasets in addition to pre-trained models to bootstrap their development \citep{devlin2018bert, brown2020language}.  A number of common techniques for improving model performance, particularly in specialized tasks, involve composing multiple models or doing some form of model fine-tuning. For example, many vision language models (VLMs) are the result of a number of existing auto-encoders, classifiers and embedding models which have their own complex provenance graph (see \cref{fig:complex_provenance} for example) \citep{moor2023medflamingo}. 

Unfortunately, many technical reports (such as model cards and system cards \citep{mitchell2019model,gursoy2022system}) do not report upstream dependencies in a format which can be easily interrogated. The consequence is that the community of researchers, system developers and users can inherit risks from the upstream models long after a foundation model is published. To address these risks, our contributions are as follows. 
\begin{itemize}
    \item We illustrate model provenance risk in the healthcare domain.
    \item We identify the properties needed to create early model warning systems.
    \item Finally, we propose an open-source, community-led system for tracking model provenance.
\end{itemize}

\subsection{Data Poisoning and Regulatory Risks}

Foundation models are defined by key artifacts: code, training and retrieval datasets, and final model parameters. Mostly for foundation models, only the model parameters are made available, while the code and datasets often remain undisclosed. For instance, ChatGPT \citep{openai2024gpt4} only provides API access, while models like Gemma \citep{google2024gemma} release parameters without sharing training details.

Deep learning models are prone to memorizing training data, leading to privacy concerns \citep{carlini2023quantifying}, even when accessed through public interfaces (i.e. in a black-box setting) \citep{shokri2017membership}. This highlights the importance of managing training artifacts and ensuring transparency in the evaluation and post-training methods like RLHF \citep{ouyang2022rlhf}. Ideally, models should include reproducible evaluation protocols to promote accountability and ethical use.

Surprisingly, even models used to embed multi-modal data--such as CLIP--can be attacked to reveal training details \citep{meehan2023dejavu}. As models are trained on internet-scale datasets, it becomes increasingly difficult to filter out problematic content. Furthermore, there are now well-known datasets, such as Common Crawl \citep{luccioni2021s}, and LAION-5B \citep{schuhmann2022laion} which have been used without any pre-processing to produce models that risk reproducing undesirable samples \citep{schuhmann2022laion}. The solution that we propose complements a growing ecosystem of related efforts described in the following section. 

\section{Related Work}
The ecosystem for model and dataset specification, model benchmarking and open efforts are already actively engaged in developing tools to help increase transparency and surface potentially undesirable behaviour during model deployment. The \textit{unified model record} we propose complements a number of existing initiatives.

\paragraph{Model and dataset specifications.}
 Model cards \citep{mitchell2019model} are designed to summarize a model's intended use, performance characteristics, ethical considerations, and other relevant details in a structured format. However, they lack the capability to model upstream relationships in a machine-readable way. Datasheets focus on how a dataset was collected while remaining agnostic to their application in different model training pipelines \citep{gebru2021datasheets}. More recently, Croissant \citep{akhtar2024croissant} has closed this gap by providing a machine-readable format for defining datasets for machine learning research but stopped short of capturing a metadata description of the models themselves. System cards \citep{gursoy2022system} have attempted to capture a more holistic description of a machine learning model and its deployment environment, and more recently, the ecosystem graph \citep{bommasani2023ecosystem} was developed to try and relate models to their upstream dependencies. These efforts are closest to our proposal. Many of these efforts have been partially adopted by some model publishers, however, they are often under-specified or vulgarized in practice \citep{liang2022holistic}. For example, at the time of this writing, Hugging Face renders a free-text README file in markdown format, denoted as a model card. More recently, they have begun tracking downstream models in a limited manner called model trees.

\paragraph{Benchmarking platforms.}
Government agencies have also begun establishing regulatory bodies focused on AI safety\citep{tabassi2023artificial}. To aid these initiatives, several benchmark datasets targeting issues such as bias \citep{vidgen2024introducingv05aisafety}, toxicity \citep{sun2024trustllm}, psychological harm \citep{kirk2024benefits}, privacy \citep{staab2023beyond} and safety \citep{kirk2024prism,bianchi2023safety} have been developed. For instance, the ML Commons group is working on an open model evaluation platform, inspired by the  HELM \citep{liang2022holistic} and HEIM \citep{lee2024holistic} frameworks \citep{vidgen2024introducingv05aisafety}. However, these benchmarking platforms require significant computational resources and currently assess only a limited number of models \citep{liang2022holistic}. These efforts build upon earlier projects like the LMSYS Chatbot Arena \citep{chiang2024chatbotarenaopenplatform}, which pioneered holistic model evaluation.

\paragraph{Open source efforts.}
On one hand, good science requires reproducible experiments, and it is clear that open-source benchmarks, models and code improve reproducibility \citep{pineau2021improving}. Reproducibility checklists and standardized disclosures further aid in evaluating scientific claims. On the other hand, while web platforms like Papers with Code, Kaggle, and Hugging Face provide essential infrastructure, they do not enforce standardization across scientific contributions.

\section{A Case Study in Healthcare}

\paragraph{Results.} We identify at least four widely published healthcare ML models that rely on upstream assets which may have been compromised. The risks identified come from the Pathology Language-Image Pertaining (PLIP) \cite{huang2023plip} model, a vision-language model widely used in the medical imaging field. We found that R2T-MIL\citep{tang2024r2t}, Breast Cancer Tumor and Immune Phenotypes Predictor \citep{goncalves2024breastcancer}, VLM-CPL \citep{zhong2024vlm-cpl}, and PathLDM \citep{yellapragada2023pathldm} each use PLIP. These vulnerabilities stem from two primary issues. First, PLIP potentially included sensitive and unethical content. Second, PLIP was trained on Twitter (now X) data, which recently has altered their data usage policies \citep{twitter_cikm_2010, qazi2020geocov19}. This may have caused any downstream models to suffer from the same legal risks as PLIP itself.

Both these cases highlight a temporal risk: researchers can publish work ethically only to find that their upstream dependencies are now considered unsuitable. Proper model provenance management would add transparency and enable early warning systems for legal, CSAM and other risks.

\paragraph{Approach.} Once a poisoned model was identified, each downstream model required at least two hours to review all available information. The whole process took in excess of 60 hours. We scrutinized popular datasets used in medical imaging, focusing on those known to have ethical concerns or potential contamination. 

\paragraph{Examining downstream dependencies for vulnerabilities.}
Having found PLIP, we then began a new search for research papers, training data documentation, and model/system cards (where available) which cited its use. This process involved carefully reading through numerous papers and technical reports, reported in a variety of formats, and with varying degrees of detail regarding model architectures and training procedures.

\begin{figure}[ht]
    \centering
\includegraphics[width=0.3\textwidth]{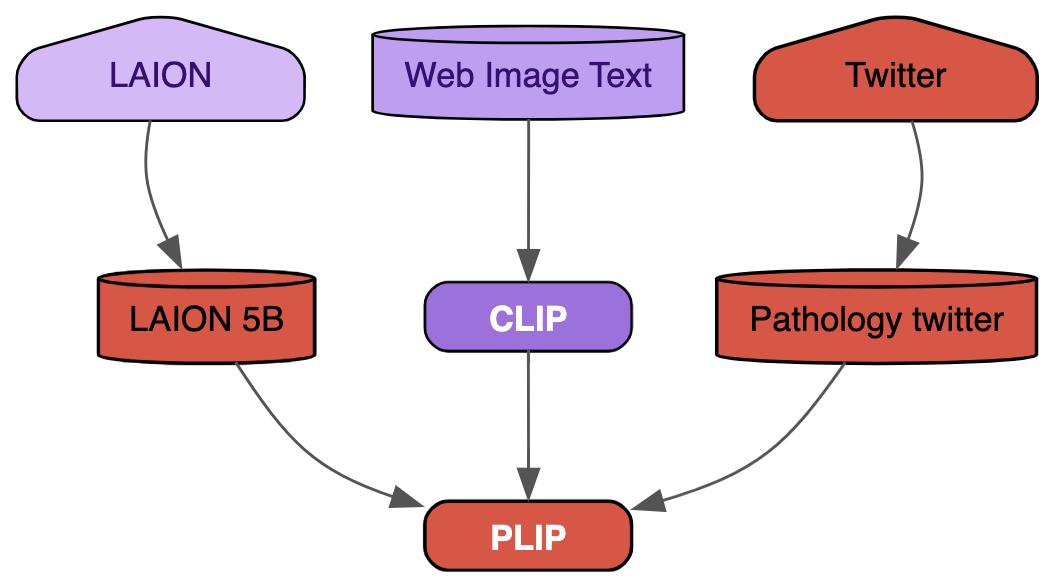}
    \caption{Model provenance graph of PLIP with its upstream dependencies}
    \label{fig:Poisoned artifact inherit to downstream models}
\end{figure}
\subsection{Threats and Implications}
\paragraph{Sensitive images in large datasets.}The use of LAION-5B \citep{schuhmann2022laion} in PLIP’s training poses significant risks. In December 2023, the Stanford Internet Observatory discovered that LAION-5B contained hundreds of known child sexual abuse material (CSAM) images \footnote{See \href{https://cyber.fsi.stanford.edu/news/investigation-finds-ai-image-generation-models-trained-child-abuse}{Stanford Internet Observatory (SIO): identified hundreds of known images of CSAM in LAION-5B}}. Although PLIP developers used only a subset of 32,000 images, they failed to rigorously filter out CSAM, meaning it may have been part of the training data. Research indicates that models like PLIP, which do not generate new images, can still retain and potentially reconstruct training samples, propagating unethical and illegal content to downstream models \citep{meehan2023dejavu}. Beyond CSAM, datasets like DukeMTMC \citep{zhang2017mtmc}, Labeled Faces in the Wild (LFW) \citep{huang2007lfw}, and MS-Celeb-1M \citep{guo2016celeb} have encountered ethical issues, such as privacy violations and lack of informed consent \citep{peng2021face}. Despite these problems and some data retractions, their downstream byproducts remain online. For instance, the LFW dataset, despite its flaws, saw over 3,000 downloads last month on platforms like Hugging Face.

\paragraph{Evolving data usage policies.}
The use of social media data, especially from platforms like X, complicates AI model development due to shifting policies. Over time, access to social media datasets has tightened -- earlier datasets, such as the 2009-2010 Cheng-Caverlee-Lee Twitter Scrape \citep{twitter_cikm_2010}, openly shared user geolocation data with full-text, while current datasets like GeoCoV19Tweets \citep{qazi2020geocov19} limit access to tweet IDs, requiring users to retrieve full-text through the X API Tools. A key concern for researchers using PLIP is whether they must comply with updated licensing policies. Do these derivative models obtain the necessary licensing? Do they recognize the potential risks of using Twitter data and verify current privacy policies upon publication? These questions remain hotly contested by lawyers and regulators \citep{nytimes_openai_microsoft_2023}. These findings stress the urgent need for better tracking of 
model and dataset provenance. Tools like our proposed \textit{unified model records} can systematically track these model dependencies. Implementing robust provenance management will improve transparency, guide better model selection, and reduce risks tied to compromised or ethically questionable datasets.

\section{Properties of Unified Model Records}
\paragraph{Community support.} 
To be successful \textit{unified model records} requires a large amount of information about models to be curated, managed and shared in an accessible manner. Such problems are more effectively addressed using open-source communities given there is joint value derived from contributing to a common repository that helps the ecosystem in a non-extractive manner. UMRs are accordingly following this model, similar to what we find in other package management systems such as PyPI or NPM. 

\paragraph{Web accessibility.}
Providing a singular source of truth to validate UMRs will also be crucial. While we can aggregate and index information about models in a distributed manner, a central index will need to be maintained to ensure data integrity across the ecosystem. This is similar to how NPM or other package managers need to be served by one or more providers, generally with one core provider, that can be augmented on a per-project or repository basis. UMR will follow a similar pattern and will provide a hosted location for model data.

\paragraph{Support for privately hosted unified model repositories.}
Much like closed-source software, private models and datasets may be produced which rely on public upstream assets. Many companies or initiatives will have needs that require managing internal-only model information. A solution exists in the open source community where \textit{private packages}\footnote{See \href{https://docs.npmjs.com/auditing-package-dependencies-for-security-vulnerabilities}{Python: Installing private packages}.} can coexist with open source efforts. Similarly, UMR provides extensibility to enable the hosting of private model data that can be merged with the publicly managed set of UMR data. This will be critical in understanding and minimizing ecosystem risks.

\paragraph{Early warning and disclosures.}
Such an approach can leverage the model provenance graph to highlight impacted model owners when a given model is known to present a risk or a change in licensing or legal frameworks makes model usage no longer tenable in certain circumstances. This is similar to how a package management can be rolled back or audit a given project's dependencies for risks on an ongoing basis\footnote{ See \href{https://docs.npmjs.com/auditing-package-dependencies-for-security-vulnerabilities}{NPM: Auditing package dependencies for security vulnerabilities}.}. UMR will enable similar abilities to highlight and detect changing issues in a given model's dependency graph and enable automated early warning to address potential risks. This becomes particularly critical as model dependencies grow in size and complexity, and managing this manually ceases to be tenable.

   

\begin{figure}[ht]
    \centering
\includegraphics[width=0.9\textwidth]{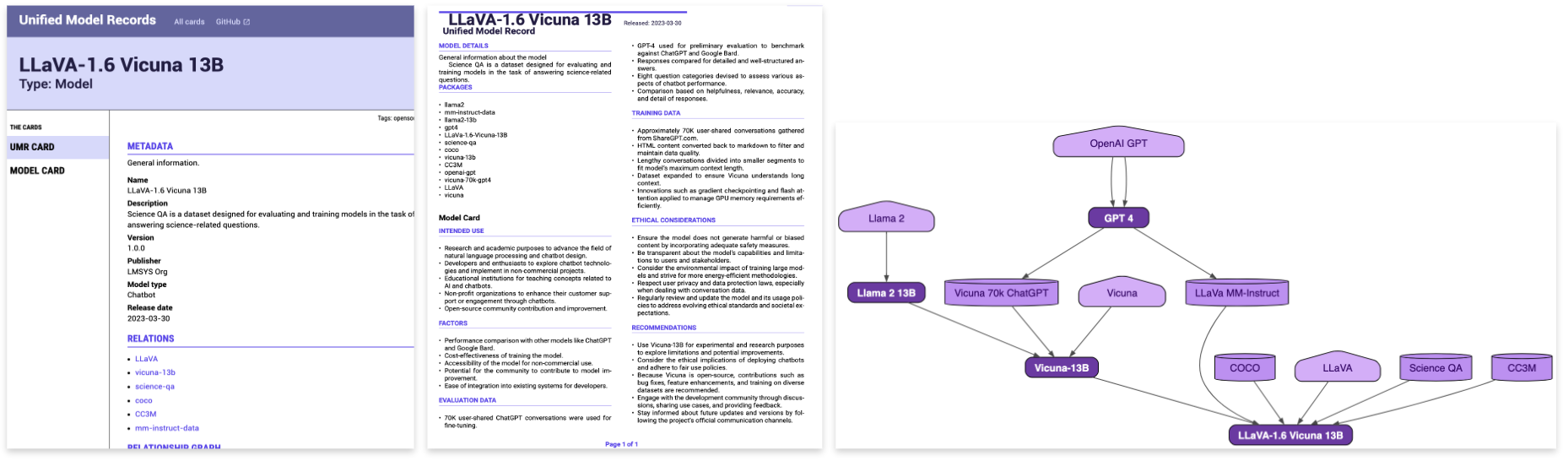}
    \caption{LLaVA-1.6 Vicuna 13B in different formats including HTML, PDF, and GraphViz}
    \label{fig: diff formats}
\end{figure}

\paragraph{Hierarchical dependency management.}
To determine the provenance graph for downstream models, we claim the following properties are required. First, all the metadata must be in standardized, machine and human-readable format, such as JSON or YAML. Second, the work of developing a consistent representation needs to be undertaken by a community of contributors. Fortunately, a successful analog exists in open-source package management systems, like PyPI or NPM, where maintainers-who may or may not be the original authors-ensure metadata integrity and track dependencies. Third, metadata must be semantically versioned to adjust downstream dependencies over time. Fourth, the system must include both qualitative and quantitative evaluation results, enabling standardized reporting across models, systems, and datasets.

\paragraph{Standardized reporting.}
Standardized reporting for foundation models and their inputs will increase the overall legibility of each artifact. Much like nutrition labels, standardized reports can speed up analysis and evaluation. For example, conference reviewers will quickly be able to compare models developed as part of academic submissions, where it would be easier to compare whether the success of a new method comes from algorithmic innovation or merely the fine-tuning of larger and larger upstream models. These model records should be made available in LaTeX, HTML, PDF and GraphViz format so that they can be readily shared. Finally, industrial labs may be incentivized to improve the metadata quality of upstream open-source contributions so that they may inherit high-quality metadata in their own model records. Currently, over 50 unified model records are available on a public website\footnote{See \href{https://www.modelrecord.com}{www.modelrecord.com}.} and Github repository \footnote{See \href{https://github.com/modelrecords/modelrecords}{Unified model records Github repository.}}.

\section{Conclusion}
The rapid advancement of foundation models necessitates robust tools for managing model provenance and mitigating downstream risks. The \textit{unified model record} (UMR) repository will serve as an index, enabling efficient tracking of model lineage and dependencies. This system will automatically notify relevant stakeholders when upstream models are flagged for potential issues, facilitating rapid response to emerging risks.

We envision integration with platforms like Hugging Face and Kaggle, where UMRs can be automatically generated and updated alongside model uploads. Future work will focus on refining the UMR format, expanding the central repository's capabilities, and collaborating with platforms to support adoption. By providing a comprehensive solution for model provenance management, UMRs aims to foster responsible innovation and mitigate risks in the rapidly evolving landscape of AI ecosystems.

\begin{ack}
Special thanks to Adriana Romero-Soriano, Ahrav Dutta, Carole-Jean W, \& Koustuv Sinha for their insightful comments and suggestions. This research was made possible by the scholarship from NSERC (Natural Sciences and Engineering Research Council of Canada).
\end{ack}

\bibliographystyle{assets/plainnat}
\bibliography{main}





\end{document}